\documentclass{article}

\usepackage{graphicx,graphics}
\usepackage[utf8]{inputenc} 
\usepackage[T1]{fontenc}    
\usepackage{hyperref}       
\usepackage{url}            
\usepackage{booktabs}       
\usepackage{amsfonts}       
\usepackage{nicefrac}       
\usepackage{microtype}      
\usepackage{mathrsfs,amsmath}
\usepackage{amssymb}
\usepackage{paralist}
\usepackage{tcolorbox}
\usepackage{pifont}
\usepackage{bbm}
\usepackage{bm}
\usepackage{multirow}
\usepackage[noend]{algpseudocode}  
\usepackage{algorithm}
\usepackage{algorithmicx}
\usepackage{xspace}
\usepackage[capitalize,noabbrev]{cleveref}
\usepackage{caption}

\usepackage{xcolor} 

\newtheorem{theorem}{Theorem}[section]

\newtheorem{corollary}[theorem]{Corollary}

\newtheorem{assumption}{\textbf{Assumption}\hspace{-2pt}}
\crefformat{assumption}{{\textbf{Assumption}}#2#1#3}

\usepackage[preprint]{cpal_2025}

\definecolor{darkgreen}{RGB}{0,128,0}

\newcommand{\ourframework}{\texttt{HASSLE-free}\xspace}

\newcommand{\Pone}{\textbf{(P1)}\xspace}
\newcommand{\Ptwo}{\textbf{(P2)}\xspace}

\definecolor{aurometalsaurus}{rgb}{0.43, 0.5, 0.5}
\definecolor{britishracinggreen}{rgb}{0.0, 0.26, 0.15}
\definecolor{burntumber}{rgb}{0.54, 0.2, 0.14}
\definecolor{cobalt}{rgb}{0.0, 0.28, 0.67}
\definecolor{bulgarianrose}{rgb}{0.28, 0.02, 0.03}
\definecolor{ceruleanblue}{rgb}{0.16, 0.32, 0.75}

\newcommand{\Nin}{N_{\text{in}}}
\newcommand{\Nout}{N_{\text{out}}}
\newcommand{\bfU}{\mathbf{U}}
\newcommand{\bfV}{\mathbf{V}}
\newcommand{\bfUVt}{\mathbf{UV}^\top}
\newcommand{\bfD}{\mathbf{D}}

\newcommand{\bfM}{\mathbf{M}}

\newcommand{\bfH}{\mathbf{H}}
\newcommand{\bfI}{\mathbf{I}}
\newcommand{\bfXTX}{\mathbf{X}^\top\mathbf{X}}
\newcommand{\bfX}{\mathbf{X}}
\newcommand{\bfZ}{\mathbf{Z}}
\newcommand{\bfW}{\mathbf{W}}
\newcommand{\bfWS}{\mathbf{W_S}}
\newcommand{\pruningW}{\tilde{\bfW}}
\newcommand{\lowrankW}{\Bar{\bfW}}
\newcommand{\bfWold}{\widehat{\mathbf{W}}}
\newcommand{\calCS}{\mathcal{C}_S}
\newcommand{\rk}[1]{\operatorname{rk}\left({#1}\right)}
\newcommand{\rkn}[1]{\operatorname{rk}({#1})}
\newcommand{\diag}[1]{\operatorname{diag}\left({#1}\right)}
\newcommand{\diagn}[1]{\operatorname{diag}({#1})}
\newcommand{\Tr}[1]{\operatorname{Tr}\left({#1}\right)}

\hypersetup{colorlinks = true, urlcolor = blue, bookmarksopen = true}

\newcommand{\N}{\mathbb{N}}

\newcommand{\R}{\mathbb{R}}

\newcommand{\argmin}{\operatornamewithlimits{\arg\min}}

\newcommand{\pr}[1]{\left({#1}\right)}

\newcommand{\prn}[1]{({#1})}

\newcommand{\norm}[1]{\left\|{#1}\right\|}

\newcommand{\abs}[1]{\left\lvert{#1}\right\rvert}

\begin{document}

\title{
\ourframework: A unified Framework for Sparse plus Low-Rank Matrix Decomposition for LLMs
}

\author{%
  Mehdi Makni\textsuperscript{1}\thanks{Corresponding author.}, ~Kayhan Behdin, ~Zheng Xu\textsuperscript{2}, ~Natalia Ponomareva\textsuperscript{2},
  ~Rahul Mazumder\textsuperscript{1}\\
  \textsuperscript{1}MIT Operations Research Center,
  \textsuperscript{2}Google Research\\
  \texttt{mmakni@mit.edu}
}

\maketitle


\begin{abstract}
The impressive capabilities of large foundation models come at a cost of substantial computing resources to serve them. Compressing these pre-trained models is of practical interest as it can democratize deploying them to the machine learning community at large by lowering the costs associated with inference.
A promising compression scheme is to decompose foundation models' dense weights into a sum of sparse plus low-rank matrices.
In this paper, we design a unified framework coined \ourframework for (semi-structured) sparse plus low-rank matrix decomposition of foundation models.
Our framework introduces the local layer-wise reconstruction error objective for this decomposition, we demonstrate that prior work solves a relaxation of this optimization problem; and we provide efficient and scalable methods to minimize the \textit{exact} introduced optimization problem. 
\ourframework substantially outperforms state-of-the-art methods in terms of the introduced objective and a wide range of LLM evaluation benchmarks. For the Llama3-8B model with a 2:4 sparsity component plus a 64-rank component decomposition, a compression scheme for which recent work shows important inference acceleration on GPUs, \ourframework reduces the test perplexity by $12\%$ for the WikiText-2 dataset and reduces the gap (compared to the dense model) of the average of eight popular zero-shot tasks by $15\%$ compared to existing methods.
\end{abstract}

\vspace{-5pt}
\section{Introduction}
\label{sec:introduction}
\vspace{-5pt}

Large Language Models (LLMs) have shown remarkable capabilities on numerous tasks in Natural Language Processing (NLP), 
ranging from language understanding to generation \cite{bubeck2023sparks, achiam2023gpt,team2023gemini, dubey2024llama}. The huge success of LLMs comes with important challenges to deploy them due to their massive size and computational costs. For instance,  Llama-3-405B \cite{dubey2024llama} requires 780GB of storage in half precision (FP16) and hence multiple high-end GPUs are needed just for inference. \textit{Model compression} has emerged as an important line of research to reduce the costs associated with deploying these foundation models. In particular, neural network pruning \cite{obd, hassibi1992second, benbaki2023fast}, where model weights are made to be sparse after training, has garnered significant attention. Different sparsity structures (Structured, Semi-Structured and Unstructured) obtained after neural network pruning result in different acceleration schemes. \textit{Structured pruning} removes entire structures such as channels, filters, or attention heads \cite{lebedev2016fast,wen2016learning,voita2019analyzing,el2022data} and readily results in acceleration as model weights dimensions are reduced. \textit{Semi-Structured pruning}, also known as, N:M sparsity \cite{zhou2021learning} requires that at most $N$ out of $M$ consecutive elements are non-zero elements. Modern NVIDIA GPUs provide support for 2:4 sparsity acceleration. \textit{Unstructured pruning} removes individual weights \cite{han2015learning, guo2016dynamic} from the model's weights and requires specialized hardware for acceleration. For instance, DeepSparse \cite{kurtic2022optimal, pmlr-v119-kurtz20a, DBLP:journals/corr/abs-2111-13445} provide CPU inference acceleration for unstructured sparsity.\\
Specializing to LLMs, one-shot pruning~\cite{meng2024alps, frantar2023sparsegpt, sun2023simple, zhang2023dynamic}, where one does a single forward pass on a small amount of calibration data, and prunes the model without expensive fine-tuning/retraining, is of particular interest. This setup requires less hardware requirements. For instance, \citet{meng2024alps} show how to prune an OPT-30B \cite{opt} using a single consumer-level V100 GPU with 32GB of CUDA memory, whereas full fine-tuning of such model using Adam \cite{kingma2014adam} at half-precision requires more than 220GB of CUDA memory.

Although one-shot pruning has desirable computational properties, it can degrade models' predictive and generative performance. To this end, recent work has studied extensions of model pruning to achieve smaller utility drop of model performance from compression. 

An interesting compression mechanism in the field of \textit{model compression} is the Sparse plus Low-Rank Matrix-Decomposition problem which aims to approximate model's weights by a sparse component plus a low-rank component~\cite{hintermuller2015robust, candes2011robust, lin2011linearized, 5394889, zhou2011godec, JMLR:v24:21-1130, NIPS2014_443cb001, yu2017compressing, li2023losparse}. Specializing to LLMs,~\citet{zhang2024oats} propose OATS 
that outperforms pruning methods for the same compression ratio (number of non-zero elements) on a wide range of LLM evaluation benchmarks (e.g. perplexity in Language generation). 

OATS \cite{zhang2024oats} is however a matrix decomposition algorithm inspired from a pruning algorithm Wanda \cite{sun2023simple}. Wanda has been designed as a relaxation/approximation of another state-of-the-art pruning algorithm SparseGPT \cite{frantar2023sparsegpt}. While Wanda has been found to be extremely useful and efficient in practice, recent work \cite{meng2024alps} show results where Wanda fails for high-sparsity regimes. In this paper, we provide a unified optimization framework to decompose pre-trained model weights into sparse plus low-rank components based on a layer-wise loss function. Our framework is modular and can incorporate different pruning and matrix-decomposition algorithms (developed independently in different contexts).
Similar to~\cite{meng2024alps} we observe that our optimization-based framework results in models with better model utility-compression tradeoffs. The difference is particularly pronounced for higher compression regimes. 

Concurrently, in a different and complementary line of work,~\citet{mozaffari2024slope} have open-sourced highly-specialized CUDA kernels designed for N:M sparse \cite{zhou2021learning} plus low-rank matrix decompositions that result in significant acceleration and memory reduction for the pre-training of LLMs.
We note that our focus here is on improved algorithms for one-shot sparse plus low-rank matrix decompositions for foundation models with billions of parameters which is different from the work of \citet{mozaffari2024slope} that focuses on accelerating the pre-training of LLMs. The designed CUDA kernels \cite{mozaffari2024slope} can be exploited in our setting for faster acceleration and reduced memory footprint during inference.


\textbf{Summary of approach.\,\,\,\,} Our framework is coined \ourframework: \underline{H}ardware-\underline{A}ware (Semi-\underline{S}tructured) \underline{S}parse plus \underline{L}ow-rank \underline{E}fficient \& approximation-\underline{free} matrix decomposition for foundation models.

Hardware-aware refers to the fact that we mostly focus on a N:M sparse \cite{zhou2021learning} plus low-rank decomposition, for which acceleration on GPUs is possible, although \ourframework supports any type of sparsity pattern (unstructured, semi-structured, structured) in the sparsity constraint. Approximation-free refers to the fact that we directly minimize the local layer-wise reconstruction error introduced in \cref{eq:matrix-decomposition}, whereas we show prior work minimizes an approximation of this objective.

We formulate the compression/decomposition task as a clear optimization problem; we minimize a local layer-wise reconstruction objective where the weights are given by the sum of a sparse and low-rank component. 
We propose an efficient Alternating-Minimization approach that scales to models with billions of parameters relying on 
two key components: one involving sparse minimization (weight sparsity) and the other involving a low-rank optimization. 
For each of these subproblems 
we discuss how approximations to the optimization task can retrieve prior algorithms.


We note that \ourframework~differs from prior one-shot (sparse) pruning methods~\cite{frantar2023sparsegpt, meng2024alps, benbaki2023fast} as we seek a sparse plus low-rank decompositon of weights.
Additionally, it differs from prior one-shot sparse plus low-rank matrix decomposition methods~\cite{zhang2024oats}
as we directly minimize the local layer-wise reconstruction objective introduced in \cref{eq:matrix-decomposition}.

Our main \textbf{contributions} can be summarized as follows.
\begin{compactitem}
    \item We introduce \ourframework a unified one-shot LLM compression framework that scales to models with billions of parameters where we directly minimize the local layer-wise reconstruction error subject to  a sparse plus low-rank matrix decomposition of the pre-trained dense weights. 

    \item \ourframework uses an Alternating-Minimization approach that iteratively minimizes a Sparse and a Low-Rank component. \ourframework uses a given pruning method as a plug-in for the subproblem pertaining to the sparse component. Additionally, it uses Gradient-Descent type methods for the subproblem pertaining to the Low-Rank component.
    
    
    

    \item We discuss how special cases of our framework relying on specific approximations of the objective retrieve popular methods such as OATS, Wanda and MP --- \cite{zhang2024oats, sun2023simple,han2015learning, sze2020efficient}. This provides valuable insights into the underlying connections across different methods. 

    \item \ourframework improves upon state-of-the-art methods for one-shot sparse plus low-rank matrix decomposition. 
    For the Llama3-8B model with a 2:4 sparsity component plus a 64-rank component decomposition, \ourframework reduces the test perplexity by $12\%$ for the WikiText-2 dataset and reduces the gap (compared to the dense model) of the average of eight popular zero-shot tasks by $15\%$ compared to existing methods.
\end{compactitem}

\vspace{-5pt}
\section{Related Work}
\label{sec:related-work}
\vspace{-5pt}

\textbf{Network pruning.\,\,\,\,} Network pruning is a well-established technique for reducing the complexity of deep neural networks by removing redundant weights \cite{obd,han2015learning}. Pruning methods can be classified based on the structure of the resulting sparse network. In terms of structure, pruning can be categorized into unstructured pruning, semi-structured pruning, and structured pruning. Unstructured pruning offers better flexibility and higher sparsity levels but requires specialized hardware for acceleration, while structured pruning is more hardware-friendly but may suffer from larger performance degradation. 
Semi-structured sparsity combines benefits of unstructured sparsity in terms of retaining the model's performance thanks to its flexibility and the benefits of structured sparsity in terms of efficiency. For instance, NVIDIA has recently introduced sparse tensor cores \cite{mozaffari2024slope} to their hardware that accelerate Gemm with N:M sparsity on modern NVIDIA GPUs. In this paper, we mostly consider N:M sparsity \cite{zhou2021learning} for the sparsity constraint, although \ourframework supports other sparsity structures.

\textbf{Sparse plus Low-Rank Matrix Decomposition.\,\,\,\,} Decomposing a weight matrix into a low-rank matrix plus a sparse matrix (also known as robust PCA) is a well-studied problem from both theoretical and algorithmic perspectives \cite{hintermuller2015robust, candes2011robust, lin2011linearized, 5394889, zhou2011godec}. These approaches have been explored by \citet{JMLR:v24:21-1130, NIPS2014_443cb001, yu2017compressing} in the context of deep learning. They have been extended to LLMs by \citet{li2023losparse} in the context of improving the utility of fine-tuning LLMs and by \citet{zhang2024oats} for sparse plus low-rank model compression.

\textbf{One-shot matrix decompositions in LLMs.\,\,\,\,} Matrix decomposition in the context of LLMs has gathered a lot of attention recently. \citet{li2023loftq, guo2023lq} decompose models' weights into a quantized weight plus a low-rank component. \citet{li2023loftq} solve this problem using an alternating-minimization approach in a data-free fashion (without using a calibration dataset). \citet{guo2023lq} consider both a data-free and a data-aware decomposition for the quantized plus low-rank decomposition problem. Their data-aware decomposition relies on an approximation of the Fisher importance matrix. \citet{zhang2024oats} develop OATS and consider a sparse plus low-rank decomposition of model's weights, and they take inspiration from the pruning algorithm Wanda \cite{sun2023simple} to incorporate outlier information (from a calibration dataset) in their decomposition. 
This paper builds on top of OATS \cite{zhang2024oats} to design an algorithm that incorporates more information from the calibration dataset in the decomposition.

\vspace{-5pt}
\section{Problem Formulation}
\label{sec:optimization-formulation}
\vspace{-5pt}

Before discussing our method, let us briefly introduce some important notations that we will use throughout the paper.

\textbf{Notation.\,\,\,\,} For a general matrix $\bfZ \in \R^{m \times n}$, 
$\rk{\bfZ}$ denotes the rank of $\bfZ$.
For a given rank $r \in \N$, $C_r(\mathbf{Z}) = \bfU_r \mathbf{\Sigma}_r \bfV_r^T$, corresponding to the matrices formed by retaining only the top-$r$ singular vectors and singular values from the full SVD of $\mathbf{Z}$. The \citet{eckart1936approximation} theorem shows that $C_r(\bfZ) = \argmin_\bfM \norm{\bfZ - \bfM},\,\, \rk{\bfM} \leq r.$

For a square matrix $\bfZ \in \R^{n \times n}$,
$\Tr{\bfZ} = \sum_{i \in [n]}\bfZ_{ii}$ denotes the trace of $\bfZ$,
$\diag{\bfZ}$ denotes the diagonal matrix $\bfD \in \R^{n \times n}$ such that $\bfD_{ii} = \bfZ_{ii}$ for any $i \in [n]$, and $\bfD_{ij} = 0$ for any $i \neq j, \, i,j \in [n]$.
We also note $\mathbf{1}_n$ and $\mathbf{0}_n$ the vector of entries all ones and all zeros respectively of size $n \in \N$.

\textbf{Layer-wise Reconstruction Error.\,\,\,\,} 
A common approach in post-training LLM compression is to decompose the full-model compression problem into layer-wise subproblems. The quality of the solution for each subproblem is assessed by measuring the $\ell_2$ error between the output of the dense layer and that of the compressed one, given a set of input activations.\\
More formally, let $\bfWold \in \R^{\Nin \times \Nout}$ denote the (dense) weight matrix of layer $\ell$, where $\Nin$ and $\Nout$ denote the input and output dimension of the layer, respectively. Given a set of $N$ calibration samples, the input activation matrix can be represented as $\bfX \in \R^{NL \times \Nin}$, where $L$ is the sequence length of an LLM. It corresponds to the output of the previous layer $\ell - 1$.  The goal of the matrix decomposition algorithm is to find a sum of a sparse weight matrix $\bfWS$ and a low-rank weight matrix $\bfM$ that minimizes the reconstruction error between the original and new layer outputs, while satisfying a target sparsity constraint and a low-rank constraint. The optimization problem is given by
\begin{equation}\label{eq:matrix-decomposition}
       \min\nolimits_{\bfWS, \bfM} \,\, \left\|\bfX \bfWold-\bfX \pr{\bfWS + \bfM}\right\|_F^2 ~~~~\text{s.t. } ~~~ \bfWS\in \calCS, ~~~\rk{\bfM} \leq r.
\end{equation}
where $\bfWS, \bfM \in \R^{\Nin \times \Nout}$, $\calCS$ denotes the the sparsity-pattern constraint set.

\vspace{-5pt}
\section{Algorithm Design}
\label{sec:algorithm-design}
\vspace{-5pt}

The Optimization of Problem \eqref{eq:matrix-decomposition} is challenging: we need to jointly find a support for $\bfWS$ so that it is feasible for the set $\calCS$, a subspace of dimension $r$ where the low-rank matrix $\mathbf{M}$ lies, and optimal weights, within these constrained sets, for both matrices to minimize the layerwise-reconstruction error. While such formulation relates to the Robust-PCA literature \cite{chandrasekaran2011rank, candes2011robust, hintermuller2015robust}, the size of parameters in $\bfWS$ and $\bfM$ can reach over 100 million in the LLM setting. For instance, the size of a down projection in a FFN of a Llama3-405b \cite{dubey2024llama} has more than 800 million parameters. 
Classical methods fail to be deployed at this scale, which makes designing novel algorithms that are more computationally efficient a necessity to solve the layerwise-reconstruction matrix decomposition problem \eqref{eq:matrix-decomposition}.

In this paper, we propose to optimize problem \eqref{eq:matrix-decomposition} using an Alternating-Minimization approach \cite{hintermuller2015robust, zhou2011godec}. We aim to decompose the problem into two 'friendlier' subproblems and iteratively minimize each one of them. In particular, we would like to iteratively solve, at iteration $t$, the subproblem \Pone, which pertains to the sparse component of the matrix decomposition.
\begin{align}\label{eq:pone-pruning}
       \bfWS^{(t+1)} &\in \argmin\nolimits_{\bfWS} \,\, \left\|\bfX \bfWold-\bfX \pr{\bfWS + \bfM^{(t)}}\right\|_F^2 ~~~~\text{s.t. } ~~~ \bfWS\in \calCS\\
       &= \argmin\nolimits_{\bfWS} \,\, \left\|\bfX \pruningW^{(t)} - \bfX \bfWS\right\|_F^2 ~~~~\text{s.t. } ~~~ \bfWS\in \calCS \tag{$\pruningW^{(t)} := \bfWold - \bfM^{(t)}$}.
\end{align}
The second subproblem to be solved, at iteration $t$, which pertains to the low-rank component of the matrix decomposition problem, is \Ptwo.
\begin{align}\label{eq:ptwo-low-rank}
       \bfM^{(t+1)} &\in \argmin\nolimits_{\bfM} \,\, \left\|\bfX \bfWold-\bfX \pr{\bfWS^{(t+1)} + \bfM}\right\|_F^2 ~~~~\text{s.t. } ~~~\rk{\bfM} \leq r\\
        &= \argmin\nolimits_{\bfM} \,\, \left\|\bfX \lowrankW^{(t+1)} -\bfX \bfM\right\|_F^2 ~~~~\text{s.t. } ~~~\rk{\bfM} \leq r \tag{$\lowrankW^{(t+1)} := \bfWold - \bfWS^{(t+1)}$}.
\end{align}
Before proceeding to discuss algorithms that solve different variations of \Pone and \Ptwo, and draw connections between existing methods in the literature of \textit{model compression}, we remove the dependence on the iteration $t$ and study \eqref{eq:pone-pruning} rewritten as follows.
\begin{align}\label{eq:general-pruning}
       \bfWS^\star 
       &\in \argmin\nolimits_{\bfWS} \,\, \left\|\bfX \pruningW - \bfX \bfWS\right\|_F^2 ~~~~\text{s.t. } ~~~ \bfWS\in \calCS\\
       &= \argmin\nolimits_{\bfWS} \,\, \Tr{(\pruningW - \bfWS)^\top \bfH (\pruningW - \bfWS)}~~~~\text{s.t. } ~~~ \bfWS\in \calCS \tag{$\bfH = \bfXTX$}. 
\end{align}

Similarly, we study \eqref{eq:ptwo-low-rank} rewritten as follows.
\begin{align}\label{eq:general-low-rank}
   \bfM^\star &\in \argmin\nolimits_{\bfM} \,\, \left\|\bfX \lowrankW -\bfX \bfM\right\|_F^2 ~~~~\text{s.t. } ~~~\rk{\bfM} \leq r\\
   &= \argmin\nolimits_{\bfM} \,\, \Tr{(\lowrankW - \bfM)^\top \bfH (\lowrankW - \bfM)} ~~~~\text{s.t. } ~~~\rk{\bfM} \leq r \notag{}.
\end{align}

\subsection{Minimizing Subproblem \Pone}\label{section-pone}
\vspace{-2pt}
To solve \eqref{eq:general-pruning}, one can consider multiple variations for $\bfH$, the Hessian of the local layer-wise reconstruction error.
\vspace{-3pt}
\subsubsection{Data-Free version: $\bfX = \bfI_{\Nin \times \Nin} \implies \bfH = \bfI_{\Nin \times \Nin}$} 
\vspace{-3pt}
A data-free pruning method (without a calibration dataset) considers $\bfX$ to be an identity matrix in \eqref{eq:general-pruning}. When $\bfH$ is an identity matrix, equation \eqref{eq:general-pruning} can be solved to optimality and an optimal solution is obtained with Magnitude Pruning (MP, \cite{han2015learning, sze2020efficient}) using a simple Hard-Thresholding operator on the dense weight $\pruningW$ -- keeping the largest values and setting the remaining values to zero. Note that MP can be applied to unstructured \cite{han2015learning}, semi-structured N:M sparsity \cite{zhou2021learning}, and structured pruning \cite{meng2024alps}. This accommodates most sparsity sets $\calCS$ in the pruning literature.

\subsubsection{Diagonal-approximation: $\bfH = \diagn{\bfXTX}$}\label{subsection-pruning-diagonal-approximation}
\vspace{-3pt}
An efficient way to approach problem \eqref{eq:general-pruning} is to approximate the Hessian of the local layer-wise reconstruction error by its diagonal. An optimal solution in this case, can be obtained by Hard-Thresholding $\bfD \pruningW$, where $\bfD = \sqrt{\diagn{\bfXTX}}$. Note that this approximation results in the state-of-the-art pruning algorithm Wanda \cite{sun2023simple}. In fact, the importance metric, $S_{ij}$ introduced in Wanda for each entry $\pruningW_{ij}$ reads as follows. Here $\bfX_j$ denotes the $j^{th}$ column of the input activation matrix $\bfX$.
\begin{equation}\label{eq:wanda-metric}
    S_{ij} = \abs{\pruningW_{ij}} \cdot \norm{\bfX_j}_2 = \abs{\bfD \pruningW}_{ij}. \tag{$\bfD = \sqrt{\diagn{\bfXTX}}$}
\end{equation}
\citet{sun2023simple} show impressive results with this approximation for unstructured and semi-structured sparsity. OATS \cite{zhang2024oats} is inspired by Wanda and decomposes model weights into sparse plus low-rank using Alternating-Minimization; their sparse update reduces to this approximation (diagonal of the local layer-wise objective's Hessian).
\vspace{-3pt}
\subsubsection{Full Hessian: $\bfH = \bfXTX + \lambda \bfI$}\label{full-hessian-pruning}
\vspace{-3pt}
This approach aims to directly minimize \eqref{eq:general-pruning}. \citet{frantar2023sparsegpt} are the first to use the full Hessian of the local layer-wise reconstruction objective \Pone at the scale of LLMs for pruning using approximations at the algorithmic level (as opposed to an approximation at the optimization formulation level). \citet{meng2024alps} extend this formulation using the operator splitting technique ADMM \cite{boyd2011distributed} and show impressive results for unstructured sparsity and N:M sparsity. \citet{meng2024osscar} extend the formulation for structured sparsity by leveraging combinatorial optimization techniques.

Our framework works as a plug-in method for any pruning algorithm to minimize \Pone at iteration $t$. Since we aim to minimize \cref{eq:matrix-decomposition} in an approximation-free manner, we select methods that use the entire Hessian as they tend to give better performance for high compression ratios. SparseGPT \cite{frantar2023sparsegpt} is a popular pruning method that considers the entire Hessian. In particular, for our numerical results, we use SparseGPT to minimize \Pone.
\vspace{-3pt}
\subsection{Minimizing Subproblem \Ptwo}
\vspace{-3pt}
As in the previous section \ref{section-pone}, we discuss algorithms and related work for different variations of $\bfH$.
\vspace{-3pt}
\subsubsection{Data-Free version: $\bfX = \bfI_{\Nin \times \Nin}$} 
\vspace{-3pt}
Drawing a line from the pruning literature, a data-free version is introduced that does not require a calibration dataset. In this case, a closed-form solution of the minimizer is given by the Truncated-SVD $C_r(\lowrankW)$. This corresponds to the best rank-$r$ approximation of $\lowrankW$. \citet{li2023loftq} use SVD on the full matrix during their low-rank minimization step for quantization plus low-rank matrix decomposition. \citet{guo2023lq} use a Randomized-SVD \cite{halko2011finding} approach for the same problem (quantization plus low-rank decomposition) instead of the full SVD since it is reduces runtime significantly while maintaining the minimization performance.
\vspace{-3pt}
\subsubsection{Diagonal-approximation: $\bfH = \diagn{\bfXTX}$}\label{subsection-diagonal-approximation}
\vspace{-3pt}
The diagonal approximation of $\bfH$ has been made popular in the pruning literature thanks to Wanda \cite{sun2023simple}. We analyze the minimizing \Ptwo with this approximation. Similar to \ref{subsection-pruning-diagonal-approximation}, we introduce $\bfD = \sqrt{\diagn{\bfXTX}}$ in equation \eqref{eq:general-low-rank}. Here we use the fact that $\bfD$ is symmetric.
\begin{align*}
   \bfM^\star 
   &\in \argmin\nolimits_{\bfM} \,\, \Tr{(\lowrankW - \bfM)^\top \bfD^2 (\lowrankW - \bfM)} ~~~~\text{s.t. } ~~~\rk{\bfM} \leq r\\
   &= \argmin\nolimits_{\bfM} \,\, \left\|\bfD \lowrankW -\bfD \bfM\right\|_F^2 ~~~~\text{s.t. } ~~~\rk{\bfM} \leq r.
\end{align*}

\begin{assumption}\label{ass:full-rank-diagonal}
The input activations matrix $\bfX$ satisfies $\diag{\bfXTX}$ is full-rank. Equivalently, no column of $\bfX$ is identically $\mathbf{0}_{N \cdot L}$.
\end{assumption}
\begin{theorem}\label{theorem:low-rank-closed-form-diag-approx}
    If \cref{ass:full-rank-diagonal} holds, then the closed-form minimizer of \eqref{eq:general-low-rank} is given by
    \begin{equation*}
        \bfM^\star = \bfD^{-1} C_r(\bfD \lowrankW).
    \end{equation*}
\end{theorem}
The proof of theorem \ref{theorem:low-rank-closed-form-diag-approx} is obtained by introducing the auxialiary variable $\tilde{\bfM} = \bfD \bfM$ and noting that $\rkn{\tilde{\bfM}} = \rkn{\bfM}$, when \cref{ass:full-rank-diagonal} holds.

Interestingly, OATS \cite{zhang2024oats} uses the same operation in the Low-Rank update of the Alternating-Minimization approach (for sparse plus low rank matrix decomposition). 
\begin{corollary}
OATS \cite{zhang2024oats} exactly minimizes \eqref{eq:matrix-decomposition} with a diagonal approximation of the Hessian of the local layer-wise reconstruction error, since they minimize \Pone and \Ptwo with the same diagonal approximation $\bfH = \diagn{\bfXTX}$.
\end{corollary}
\vspace{-2pt}
\subsubsection{Full Hessian: $\bfH = \bfXTX + \lambda \bfI$}
\vspace{-3pt}
The state-of-the-art pruning algorithms in terms of retaining compressed LLMs performance on multiple benchmarks are the ones that use the full Hessian \ref{full-hessian-pruning} \cite{meng2024alps,frantar2023sparsegpt}. This motivates minimizing equation \eqref{eq:general-low-rank} using the full Hessian as well.
Dealing with low-rank constraints can be challenging, we therefore propose to reparametrize the low-rank matrix $\bfM \in \R^{\Nin \times \Nout}$ by  $\bfUVt$, with $\bfU \in \R^{\Nin \times r}, \bfV \in \R^{\Nout \times r}$. We can therefore use more computationally efficient first-order optimization methods to minimize the layer-wise reconstruction objective, which can be rewritten as follows.
\begin{equation}\label{eq:uvt-general-low-rank}
    \bfM^\star = \bfU^\star \bfV^{\star^\top}, \quad \bfU^\star, \bfV^\star \in \argmin\nolimits_{\bfU, \bfV} \,\, \Tr{\pr{\lowrankW - \bfUVt}^\top \bfH \pr{\lowrankW - \bfUVt}}.
\end{equation}
\textbf{Diagonal Scaling for \Ptwo Minimization Stability.\,\,\,\,}\label{scaling-low-rank}
Our initial experiments to minimize equation \eqref{eq:uvt-general-low-rank}, using Gradient-Descent type methods on $\bfU$ and $\bfV$, have shown that the optimization problem can be ill-conditioned in some tranformer layers. This can lead to numerical instability in the optimization procedure. To address this, we follow a similar rescaling approach proposed by \citet{meng2024alps}. Define (similar to \ref{subsection-diagonal-approximation}) the matrix $\bfD = \sqrt{\diagn{\bfXTX}}$ and reformulate the optimization equation \eqref{eq:uvt-general-low-rank} as follows (when \cref{ass:full-rank-diagonal} holds).
\begin{equation}\label{eq:scaled-uvt-general-low-rank}
    \bfM^\star = \bfD\bfU^\star \bfV^{\star^\top}, \quad \bfU^\star, \bfV^\star \in \argmin\nolimits_{\bfU, \bfV} \,\, \Tr{\pr{\bfD\lowrankW - \bfUVt}^\top \bfD^{-1}\bfH\bfD^{-1} \pr{\bfD\lowrankW - \bfUVt}}.
\end{equation}
It is important to note that the minimization problems in equations \eqref{eq:scaled-uvt-general-low-rank} and \eqref{eq:uvt-general-low-rank} are equivalent, in terms of objective minimization and feasibility of $\bfM^\star$, which has rank at most $r$. This scaling, which sets the diagonal of the new Hessian to $\mathbf{1}_{\Nin}$, only modifies the steps of Gradient Descent and leads to faster convergence in practice. See the \Cref{fig:reconstruction-error} for an objective minimization comparison of the effect of this Diagonal scaling.
It is also worth noting that this scaling allows to use the same learning rate $\eta$ for Gradient-Descent type methods for all layers and all models.

\subsection{Our Proposed Approach}
\vspace{-5pt}
Our goal is to minimize \eqref{eq:matrix-decomposition} using Alternating-Minimization with the full Hessian approach $\bfH = \bfXTX$. Our numerical results show that leveraging the entire Hessian outperforms OATS \cite{zhang2024oats}, which minimizes \eqref{eq:matrix-decomposition} with the diagonal approximation of the Hessian approach $\bfH = \diagn{\bfXTX}$, on a wide-range of LLM benchmarks and compression ratios. 

In our numerical experiments, we show results with the SparseGPT \cite{frantar2023sparsegpt} algorithm to minimize \Pone and the Adam algorithm \cite{kingma2014adam} to minimize \Ptwo reparametrized and rescaled as in \cref{eq:scaled-uvt-general-low-rank}. 

\textbf{Computational Efficiency.\,\,\,\,} Note that for a given layer $\ell$, the Hessian of the local layer-wise reconstruction problem $\bfXTX$ in \eqref{eq:general-pruning} as well as the rescaled version $\bfD^{-1}\bfXTX\bfD^{-1}$ in \eqref{eq:general-low-rank} are invariant throughout iterations. 
This is very important as pruning algorithms that use the entire Hessian information \cite{frantar2023sparsegpt, meng2024alps} need the Hessian inverse in their algorithm update. This inversion and associated costs of Hessian construction are done only once and then \textit{amortized} throughout iterations. In the \cref{algo:low-rank-gd}, we use $\bfU^{(t-1)}$ and $\bfV^{(t-1)}$ as initializations for the optimizer, as they are close to the minimizers of \Ptwo at iteration $t$. This accelerates the convergence in practice.


\begin{algorithm}[h]
    \caption{\texttt{Low-Rank-GD}
    }
    \label{algo:low-rank-gd}
    \begin{algorithmic}[]
        \State \textbf{Input} \texttt{Optimizer} (optimization algorithm, e.g. Adam), $\bfH$ (Hessian), $\bfW$ (Weights), $\bfU_\text{init}$, $\bfV_\text{init}$ (warm-up initialization for the joint minimization of $\bfU, \bfV$), $T_{\text{LR}}$ (\# iterations), $\eta$ (learning rate).
        \State $\text{Obj}(\bfU, \bfV) \gets \Tr{\prn{\bfW - \bfUVt}^\top \bfH \prn{\bfW - \bfUVt}}$        \vspace*{0.4em}
        \State $\bfU^\star, \bfV^\star \gets \texttt{Optimizer}_{\bfU, \bfV}\pr{\text{Obj}, \bfU_\text{init}, \bfV_\text{init}, N, \eta}$
        \vspace*{0.4em}
        \State \textbf{Output} $\bfU^\star, \bfV^\star$.
    \end{algorithmic}
\end{algorithm}
\vspace{-\baselineskip}
\begin{algorithm}[h]
    \caption{\ourframework}
    \label{algo:gdprune}
    \begin{algorithmic}[]
        \State \textbf{Input} for a given layer $\ell$: $\mathbf{H} = (\bfXTX + \lambda \mathbf{I})$ (Hessian of \eqref{eq:matrix-decomposition}, plus a regularization term for numerical stability), $\widehat{\bfW}$ (dense pre-trained weights), $T_{\text{AM}}$ (\# iterations of Alternating-Minimization), $T_{\text{LR}}$ (\# iterations of \texttt{Low-Rank-GD}), $\eta$ (learning rate for $\bfU, \bfV$), $\calCS$ (sparsity pattern), $r$ (rank of low-rank components), \texttt{Prune} (any pruning algorithm, e.g. SparseGPT), \texttt{Optimizer} (any first-order algorithm, e.g. Adam), \textbf{is\_scaled} (bool to apply scaling \ref{scaling-low-rank}).
        \vspace*{0.1em}
        \State $\bfD \gets \sqrt{\diag{\bfH}}$ \quad \Comment{Diagonal of the Hessian.}
        \State $\mathbf{H}^{-1} \gets \texttt{inv}\pr{\mathbf{H}}$ \quad \Comment{Inverse the Hessian.}
        \State $\bfWS \gets \mathbf{0}_{\Nin \times \Nout}$
        \State $\bfU \gets \mathbf{0}_{\Nin \times r}$
        \State $\bfV \gets \mathcal{N}_{\Nout \times r}$ \quad  \Comment{element-wise independent gaussian initialization.}
        \For{$t = 1 \dots T$}
            \State $\bfWS \gets \texttt{Prune}\pr{\mathbf{H}^{-1}, \widehat{\bfW} - \bfUVt, \calCS}$
            \State \Comment{$\bfWS \approx \widehat{\bfW} - \bfUVt$, satisfies $\calCS$ sparsity pattern \& minimizes \Pone.}
            \vspace*{0.2em}
            \State $\eta_t \gets \text{get\_lr}(t, \eta)$ \quad \Comment{In practice, $\eta_t = \eta / (t + 10)$.}
            \vspace*{0.2em}
            \If{\textbf{is\_scaled}}
                \State $\bfU, \bfV \gets \texttt{Low-Rank-GD}\pr{\texttt{Optimizer}, \bfD^{-1}\mathbf{H}\bfD^{-1}, \bfD\pr{\widehat{\bfW} - \bfWS}, \bfD\bfU, \bfV, T_{\text{LR}}, \eta_t}$
                \State $\bfU \gets \bfD \bfU$ \quad \Comment{Rescale $\bfU$ back.}
            \Else
                \State $\bfU, \bfV \gets \texttt{Low-Rank-GD}\pr{\texttt{Optimizer}, \mathbf{H}, \widehat{\bfW} - \bfWS, \bfU, \bfV, T_{\text{LR}}, \eta_t}$ 
            \EndIf
            \State \Comment{$\bfUVt \approx \widehat{\bfW} - \bfWS$, has rank at most $r$ \& minimizes \Ptwo.}
        \EndFor
        \State $\bfM \gets \bfUVt$
        \State \textbf{Output} for a given layer $\ell$: $\bfWS, \bfM$.
    \end{algorithmic}
\end{algorithm}

\vspace{-5pt}
\section{Experimental Results}
\label{sec:experimental-results}
\vspace{-5pt}

\subsection{Experiment Setup}
\vspace{-3pt}
\textbf{Models and datasets} We evaluate our proposed method \ourframework on two families of large language models: Llama-3 and Llama-3.2 \cite{dubey2024llama} with sizes ranging from 1 to 8 billion parameters. 
To construct the Hessian $\bfXTX$, we follow the approch of \citet{frantar2023sparsegpt}: we use 128 segments of 2048 each, randomly sampled from the first shard of the C4 training dataset \cite{JMLR:v21:20-074}. To ensure consistency, we utilize the same calibration data for all pruning algorithms we benchmark. We also consider one-shot compression results, without retraining. 
We assess the performance using perplexity and zero-shot evaluation benchmarks, with perplexity calculated according to the procedure described by HuggingFace \cite{Perplexity}, using full stride. For perplexity evaluations, we use the test sets of raw-WikiText2 \cite{merity2017pointer}, PTB \cite{Marcus1994}, and a subset of the C4 validation data, which are popular benchmarks in LLM pruning literature \cite{frantar2023sparsegpt,meng2024alps,meng2024osscar}. Additionally, we evaluate the following zero-shot tasks using LM Harness by \citet{gao10256836framework}: PIQA \cite{bisk2020piqa}, ARC-Easy (ARC-E) \& ARC-Challenge (ARC-C) \cite{clark2018think}, Hellaswag (HS) \cite{zellers2019hellaswag}, Winogrande (WG) \cite{sakaguchi2021winogrande}, RTE \cite{poliak2020survey}, OpenbookQA (OQA) \cite{banerjee2019careful} and BoolQ \cite{clark2019boolq}. The average of the eight zero-shot tasks is also reported.

\vspace{-5pt}
\subsection{Results}
\vspace{-10pt}
In order to benchmark the performance of our matrix decomposition algorithm, \ourframework uses the same number of Alternating-Minimization steps as OATS \cite{zhang2024oats} which is $80$. We report results for the scaled version of \ourframework, which uses the same learning rate $\eta = 1e^{-2}$ for all layers and considered models. We consider the following two settings.

\textbf{N:M Sparsity + Fixed Rank:}
We impose the sparsity pattern $\calCS$ to be $N:M$ sparsity and we fix the target rank $r = 64$ of the low-rank component for all layers. We benchmark our method with OATS \cite{zhang2024oats}. The results are reported in \cref{tab:slr-fixed-rank}.

\textbf{N:M Sparsity + Fixed Compression Ratio:}
This is similar to the setting described by \citet{zhang2024oats} for N:M sparsity evaluations. Each layer, with dense weight matrix $\widehat{\bfW}$, is compressed to a prefixed compression ratio $\rho$ (e.g. $50\%$) so that $\widehat{\bfW} \approx \bfW_{N:M} + \bfM$, and the target rank is given by\\[0.2em] 
$r = \left\lfloor {(1 - \rho - \frac{N}{M}) \cdot(\Nout \cdot \Nin)} / \pr{\Nout + \Nin} \right\rfloor$.\\[0.2em]
Note that the effective number of parameters stored is therefore\\[0.1em]
$\#\text{params } \bfW_{N:M} + \#\text{params } \bfU + \#\text{params } \bfV = \frac{N}{M} \cdot (\Nout \cdot \Nin) + r \Nin + r \Nout \leq (1 - \rho) \cdot \#\text{params } \widehat{\bfW}$,\\[0.5em]
hence the comparison to other pruning methods matched at the same compression ratio $\rho$. The results are reported for the Llama3-8B model in \cref{tab:slr-fixed-compression} for \ourframework, \texttt{OATS}, and different N:M pruning algorithms (SparseGPT \cite{frantar2023sparsegpt}, Wanda \cite{sun2023simple}, DSNoT \cite{zhang2023dynamic}) compressed at $\rho = 50\%$. The results are expanded for \ourframework and OATS in \cref{supp:experiments-supp}. 


\vspace{-8pt}
\noindent
\begin{minipage}[t]{0.5\textwidth}
    \raggedleft
    \vspace{25pt}
    \captionof{table}{Performance analysis for one-shot N:M sparse plus a low-rank matrix decomposition of the Llama3-8b model. The compression ratio is fixed to be $\rho=0.5$. For Perplexity, $(\downarrow)$ lower values are preferred. For zero-shot tasks, $(\uparrow)$ higher values are preferred.\\
    Bolded values correspond to a comparison between sparse plus low-rank decomposition algorithms. Underlined values correspond to the overall best comopression scheme given a compression ratio $\rho = 50\%$.}
    \label{tab:slr-fixed-compression}
\end{minipage}%
\hspace{5pt}
\begin{minipage}[t]{0.5\textwidth}
\centering
\begin{table}[H]
\centering
\resizebox{1.0\textwidth}{!}{%
\renewcommand{\arraystretch}{1.3}
\begin{tabular}{ccccccccc}
\toprule
\multirow{2.25}{*}{\textbf{Algorithm}} && \multicolumn{3}{c}{\textbf{Perplexity ($\downarrow$)}} && \multicolumn{3}{c}{\textbf{Zero-shot ($\uparrow$)}} \\ 
\cmidrule(rl){3-5} \cmidrule(rl){7-9}
&&\textbf{C4} & \textbf{WT2} & \textbf{PTB} && \textbf{PIQA} & \textbf{ARC-E} & \textbf{ARC-C}\\ 
\midrule
\texttt{SparseGPT-4:8}     && \underline{14.94} & 12.40 & 17.90 && 73.20 & \underline{68.54} & 34.86 \\ 
\texttt{Wanda-4:8}         && 18.88 & 14.52 & 24.26 && 71.52 & 64.91 & 34.03 \\ 
\texttt{DSNoT-4:8}         && 18.89 & 14.76 & 23.90 && 71.49 & 65.65 & 33.57 \\ 
\cmidrule(rl){1-1}
\texttt{SparseGPT-2:4}     && 18.89 & 16.35 & 25.08 && 70.54 & 63.09 & 31.84 \\ 
\texttt{Wanda-2:4}         && 30.81 & 24.36 & 44.89 && 67.56 & 56.20 & 26.11 \\ 
\texttt{DSNoT-2:4}         && 28.78 & 23.09 & 40.95 && 67.70 & 56.46 & 25.68 \\
\cmidrule(rl){1-1}
\texttt{OATS-2:8+LR}       && 21.03 & \textbf{14.54} & 24.15 && 73.67  & 59.68 & \textbf{37.12}\\
\texttt{Ours-2:8+LR}       && \textbf{20.05} & 15.03 & \textbf{22.01} && \textbf{74.05} & \textbf{60.52} & 36.18\\
\cmidrule(rl){1-1}
\texttt{OATS-3:8+LR}       && 16.87 & 11.43 & 18.53 && 75.24 & 65.91 & 39.85 \\
\texttt{Ours-3:8+LR}       && \textbf{16.16} & \underline{\textbf{11.36}} & \underline{\textbf{16.71}} && \underline{\textbf{75.79}} & \textbf{67.55} & \underline{\textbf{41.04}} \\
\cmidrule(rl){1-1}
\texttt{dense}               && 9.44 & 6.14 & 11.18 && 80.79 & 77.69 & 53.33  \\
\bottomrule
\end{tabular}
}
\end{table}
\end{minipage}

\begin{table}[h!]
\centering
\resizebox{1.0\textwidth}{!}{%
\renewcommand{\arraystretch}{1.1}
\begin{tabular}{ccc@{\hskip 8pt}cccc@{\hskip 8pt}ccccccccc}
\toprule
\multirow{2.25}{*}{\textbf{Model}} & \multirow{2.25}{*}{\textbf{Algorithm}} && \multicolumn{3}{c}{\textbf{Perplexity ($\downarrow$)}} && \multicolumn{9}{c}{\textbf{Zero-shot ($\uparrow$)}} \\ 
\cmidrule(rl){4-6} \cmidrule(r){8-16}
&&& \textbf{C4} & \textbf{WT2} & \textbf{PTB} && \textbf{PIQA} & \textbf{HS} & \textbf{ARC-E} & \textbf{ARC-C} & \textbf{WG} & \textbf{RTE} & \textbf{OQA} & \textbf{BoolQ} & \textbf{Avg}\\ 
\midrule
\multirow{9.5}{*}{Llama3-8B} 
&\texttt{OATS-2:8+64LR}       && 368.24 & 416.14 & 565.46 && 52.29 & 28.03 & 27.53 & \textbf{22.70} & 49.17 & \textbf{52.71} & 26.40 & 42.08 & 37.61 \\
&\texttt{Ours-2:8+64LR}       && \textbf{90.46} & \textbf{92.59} & \textbf{108.80} && \textbf{54.52} & \textbf{30.85} & \textbf{31.44} & 20.73 & \textbf{50.20} & \textbf{52.71} & \textbf{26.60} & \textbf{60.37} & \textbf{40.93} \\
\cmidrule(rl){2-2}
&\texttt{OATS-3:8+64LR}       && 48.21 & 35.65 & 56.52 && 65.23 & 42.05 & 47.01 & 25.94 & 58.01 & 52.71 & 27.40 & 67.89 & 48.28 \\
&\texttt{Ours-3:8+64LR}       && \textbf{28.88} & \textbf{21.48} & \textbf{32.54} && \textbf{68.99} & \textbf{52.19} & \textbf{50.55} & \textbf{29.86} & \textbf{62.90} & \textbf{53.07} & \textbf{29.80} & \textbf{72.84} & \textbf{52.53} \\
\cmidrule(rl){2-2}
&\texttt{OATS-4:8+64LR}       && 15.97 & 10.52 & 16.71 && 75.14 & 68.69 & 66.67 & 40.87 & 69.69 & \textbf{54.87} & 39.40 & \textbf{79.76} & 61.89 \\
&\texttt{Ours-4:8+64LR}       && \textbf{14.67} & \textbf{9.93} & \textbf{15.28} && \textbf{76.39} & \textbf{70.48} & \textbf{68.48} & \textbf{42.58} & \textbf{70.32} & 54.15 & \textbf{39.80} & 79.48 & \textbf{62.71} \\
\cmidrule(rl){2-2}
&\texttt{OATS-2:4+64LR}       && 21.05 & 14.42 & 22.62 && 72.85 & 62.47 & 60.69 & 36.35 & 67.09 & 54.87 & 35.00 & 75.11 & 58.05 \\
&\texttt{Ours-2:4+64LR}       && \textbf{18.06} & \textbf{12.66} & \textbf{18.66} && \textbf{74.86} & \textbf{64.77} & \textbf{63.85} & \textbf{37.37} & \textbf{69.22} & \textbf{56.68} & \textbf{36.40} & \textbf{76.12} & \textbf{59.91} \\
\cmidrule(rl){2-2}
&\texttt{dense}               && 9.44 & 6.14 & 11.18 && 80.79 & 79.17 & 77.69 & 53.33 & 72.85 & 69.68 & 45.00 & 81.44 & 69.99 \\

\midrule
\multirow{9.5}{*}{Llama3.2-1B} 
&\texttt{OATS-2:8+64LR}       && 740.37 & 825.40 & 754.22 && 52.12 & 27.46 & 28.37 & \textbf{23.72} & 48.86 & 52.71 & 24.60 & 37.77 & 36.95 \\
&\texttt{Ours-2:8+64LR}       && \textbf{167.87} & \textbf{133.01} & \textbf{162.73} && \textbf{54.30} & \textbf{28.73} & \textbf{30.35} & 21.93 & \textbf{50.51} & \textbf{53.43} & \textbf{25.20} & \textbf{51.68} & \textbf{39.52} \\
\cmidrule(rl){2-2}
&\texttt{OATS-3:8+64LR}       && 96.32 & 74.10 & 93.70 && 59.52 & 33.51 & 36.41 & 22.70 & 50.99 & \textbf{52.71} & 25.80 & 62.14 & 42.97 \\
&\texttt{Ours-3:8+64LR}       && \textbf{45.79} & \textbf{34.15} & \textbf{52.20} && \textbf{62.08} & \textbf{38.24} & \textbf{41.04} & \textbf{23.63} & \textbf{54.54} & \textbf{52.71} & \textbf{30.40} & \textbf{62.20} & \textbf{45.60} \\
\cmidrule(rl){2-2}
&\texttt{OATS-4:8+64LR}       && 26.75 & 18.49 & 31.94 && 67.30 & 49.52 & 50.51 & 28.41 & 56.67 & \textbf{55.96} & \textbf{32.40} & \textbf{62.87} & 50.46 \\
&\texttt{Ours-4:8+64LR}       && \textbf{22.71} & \textbf{16.05} & \textbf{26.80} && \textbf{68.28} & \textbf{51.42} & \textbf{51.22} & \textbf{29.18} & \textbf{58.64} & 53.07 & 30.00 & 62.51 & \textbf{50.54} \\
\cmidrule(rl){2-2}
&\texttt{OATS-2:4+64LR}       && 36.89 & 26.26 & 42.35 && 64.36 & 43.35 & \textbf{47.77} & 26.45 & 55.80 & \textbf{52.71} & 30.40 & \textbf{62.66} & 47.94 \\
&\texttt{Ours-2:4+64LR}       && \textbf{27.09} & \textbf{19.57} & \textbf{31.73} && \textbf{67.03} & \textbf{47.53} & 47.43 & \textbf{28.16} & \textbf{58.64} & \textbf{52.71} & \textbf{30.60} & 62.60 & \textbf{49.34} \\
\cmidrule(rl){2-2}
&\texttt{dense}               && 14.01 & 9.75 & 17.59 && 74.59 & 63.66 & 60.48 & 36.26 & 60.69 & 56.68 & 37.20 & 63.98 & 56.69 \\

\midrule
\multirow{9.5}{*}{Llama3.2-3B} 
&\texttt{OATS-2:8+64LR}       && 444.37 & 543.53 & 851.16 && 52.56 & 27.54 & 27.99 & \textbf{23.46} & \textbf{50.43} & 51.99 & \textbf{26.60} & 37.86 & 37.30 \\
&\texttt{Ours-2:8+64LR}       && \textbf{122.14} & \textbf{114.74} & \textbf{165.78} && \textbf{54.57} & \textbf{28.93} & \textbf{30.09} & 21.08 & 49.49 & \textbf{52.71} & 26.20 & \textbf{62.14} & \textbf{40.65} \\
\cmidrule(rl){2-2}
&\texttt{OATS-3:8+64LR}       && 56.80 & 41.62 & 72.75 && 62.68 & 40.49 & 41.84 & 24.06 & 53.91 & 52.35 & 26.60 & 64.10 & 45.75 \\
&\texttt{Ours-3:8+64LR}       && \textbf{35.07} & \textbf{27.12} & \textbf{39.63} && \textbf{66.43} & \textbf{46.08} & \textbf{46.42} & \textbf{26.62} & \textbf{58.17} & \textbf{55.96} & \textbf{29.00} & \textbf{65.47} & \textbf{49.27} \\
\cmidrule(rl){2-2}
&\texttt{OATS-4:8+64LR}       && 18.52 & 12.85 & 20.69 && 72.85 & 61.68 & 62.42 & 36.01 & 64.17 & \textbf{60.29} & 36.40 & \textbf{72.75} & 58.32 \\
&\texttt{Ours-4:8+64LR}       && \textbf{17.19} & \textbf{12.15} & \textbf{19.24} && \textbf{73.99} & \textbf{63.59} & \textbf{62.92} & \textbf{36.26} & \textbf{67.48} & 57.76 & \textbf{39.20} & 71.90 & \textbf{59.14} \\
\cmidrule(rl){2-2}
&\texttt{OATS-2:4+64LR}       && 24.32 & 17.06 & 28.54 && \textbf{71.98} & 55.87 & 58.80 & 33.36 & 59.91 & 53.07 & \textbf{33.80} & \textbf{70.18} & 54.62 \\
&\texttt{Ours-2:4+64LR}       && \textbf{20.82} & \textbf{15.65} & \textbf{23.77} && 71.71 & \textbf{57.88} & \textbf{58.84} & \textbf{34.39} & \textbf{62.12} & \textbf{58.12} & 33.60 & 67.92 & \textbf{55.57} \\
\cmidrule(rl){2-2}
&\texttt{dense}               && 11.33 & 7.81 & 13.53 && 77.48 & 73.61 & 71.63 & 45.99 & 69.85 & 54.51 & 43.00 & 73.39 & 63.68 \\

\bottomrule
\end{tabular}
}
\vspace{3pt}
\caption{Performance analysis for one-shot N:M sparse plus a 64-rank low-rank matrix decomposition of Llama3 and Llama3.2 models. The rank of the low-rank component is fixed to be $r=64$. For Perplexity, $(\downarrow)$ lower values are preferred. For zero-shot tasks, $(\uparrow)$ higher values are preferred.}
\label{tab:slr-fixed-rank}
\end{table}
\subsection{Reconstruction error on a single Transformer block}
In order to show the performance of OATS and \ourframework on the layer-wise reconstruction objective \eqref{eq:matrix-decomposition}, we compute the error produced with the two algorithms (both after $80$ iterations--default value used in OATS \cite{zhang2024oats}), given by $\|\bfX \bfWold-\bfX \pr{\bfWS + \bfM}\|_F^2$, when applied to the model Llama-3-8B \cite{dubey2024llama}, and using the decomposition $\calCS$ corresponding to $2:4$ sparsity and a fixed rank $r = 64$. Results of the local layer-wisre error are reported in \cref{fig:reconstruction-error} for OATS, \ourframework scaled and \ourframework unscaled.

\vspace{-20pt}
\begin{minipage}[t]{0.49\textwidth}
    \raggedleft
    \vspace{30pt}
    \captionof{figure}{Local layer-wise reconstruction error $\downarrow$ (lower values are preferred) analysis of the decomposition of the layers of the \textbf{first} transformer block in Llama-3-8B into a 2:4 sparse component plus a 64-rank low-rank component. All methods use the same number of Alternating-Minimization steps $80$.}
    \label{fig:reconstruction-error}
\end{minipage}
\begin{minipage}[t]{0.49\textwidth}
    \vspace{0pt}
    \centering
    \includegraphics[width=\textwidth]{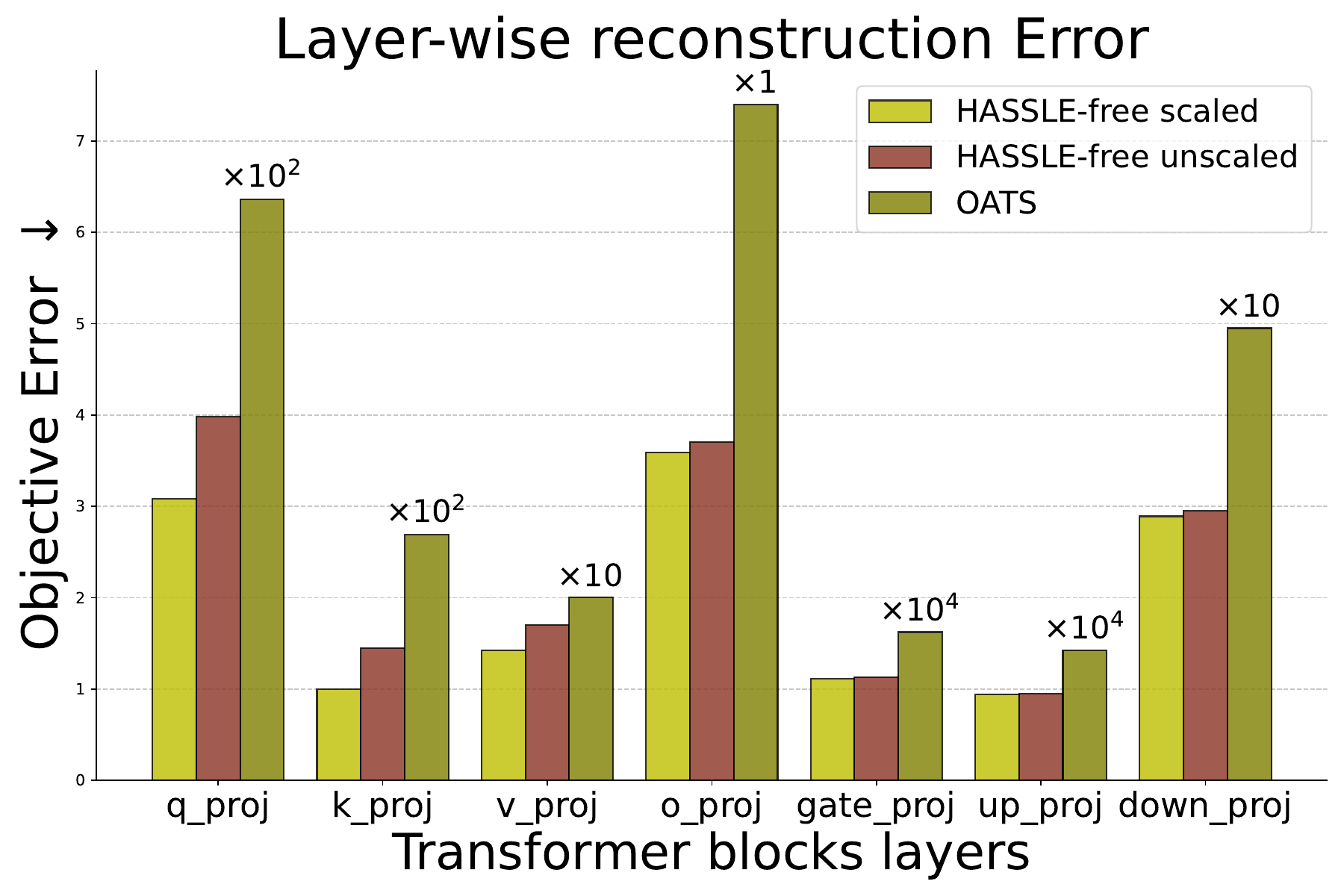}
\end{minipage}
\vspace{-\baselineskip}

\vspace{-15pt}
\section{Conclusion}
\label{sec:conclusion}
\vspace{-5pt}

We present \ourframework, a unified framework for one-shot sparse plus low-rank matrix decomposition for foundation models. \ourframework employs an Alternating Minimization approach to minimize the local layer-wise reconstruction objective without any approximations at the objective level. It scales to models with billions of parameters and it is made efficient by exploiting the problem structure (e.g. Hessian-invariance throughout iterations and diagonal rescaling of a minimization approach). Our experiments show that \ourframework outperforms existing methods for sparse plus low-rank decomposition of LLMs on a wide-range of LLM evaluation benchmarks, especially perplexity. 
Future work can extend \Pone (subproblem pertaining to sparsity) to include quantization and quantized-sparse compression. This would give a better understanding of optimization-based approaches in decomposing dense pre-trained weights into a compressed version (e.g. quantized) plus a low-rank component.

\section*{Acknowledgements}
This research is supported in part by grants from Google and the Office of Naval Research. We acknowledge the MIT SuperCloud~\cite{reuther2018interactive} for providing HPC resources that have contributed to the research results reported within this paper. We also acknowledge Google for providing us with Google Cloud Credits for computing. 

\clearpage

\bibliography{reference}

\begin{thebibliography}{55}
\providecommand{\natexlab}[1]{#1}
\providecommand{\url}[1]{\texttt{#1}}
\expandafter\ifx\csname urlstyle\endcsname\relax
  \providecommand{\doi}[1]{doi: #1}\else
  \providecommand{\doi}{doi: \begingroup \urlstyle{rm}\Url}\fi

\bibitem[Bubeck et~al.(2023)Bubeck, Chandrasekaran, Eldan, Gehrke, Horvitz,
  Kamar, Lee, Lee, Li, Lundberg, et~al.]{bubeck2023sparks}
S{\'e}bastien Bubeck, Varun Chandrasekaran, Ronen Eldan, Johannes Gehrke, Eric
  Horvitz, Ece Kamar, Peter Lee, Yin~Tat Lee, Yuanzhi Li, Scott Lundberg,
  et~al.
\newblock Sparks of artificial general intelligence: Early experiments with
  gpt-4.
\newblock \emph{arXiv preprint arXiv:2303.12712}, 2023.

\bibitem[Achiam et~al.(2023)Achiam, Adler, Agarwal, Ahmad, Akkaya, Aleman,
  Almeida, Altenschmidt, Altman, Anadkat, et~al.]{achiam2023gpt}
Josh Achiam, Steven Adler, Sandhini Agarwal, Lama Ahmad, Ilge Akkaya,
  Florencia~Leoni Aleman, Diogo Almeida, Janko Altenschmidt, Sam Altman,
  Shyamal Anadkat, et~al.
\newblock Gpt-4 technical report.
\newblock \emph{arXiv preprint arXiv:2303.08774}, 2023.

\bibitem[Google(2023)]{team2023gemini}
Gemini~Team Google.
\newblock Gemini: a family of highly capable multimodal models.
\newblock \emph{arXiv preprint arXiv:2312.11805}, 2023.

\bibitem[Dubey et~al.(2024)Dubey, Jauhri, Pandey, Kadian, Al-Dahle, Letman,
  Mathur, Schelten, Yang, Fan, et~al.]{dubey2024llama}
Abhimanyu Dubey, Abhinav Jauhri, Abhinav Pandey, Abhishek Kadian, Ahmad
  Al-Dahle, Aiesha Letman, Akhil Mathur, Alan Schelten, Amy Yang, Angela Fan,
  et~al.
\newblock The llama 3 herd of models.
\newblock \emph{arXiv preprint arXiv:2407.21783}, 2024.

\bibitem[LeCun et~al.(1989)LeCun, Denker, and Solla]{obd}
Yann LeCun, John Denker, and Sara Solla.
\newblock Optimal brain damage.
\newblock \emph{Advances in neural information processing systems}, 2, 1989.

\bibitem[Hassibi and Stork(1992)]{hassibi1992second}
Babak Hassibi and David Stork.
\newblock Second order derivatives for network pruning: Optimal brain surgeon.
\newblock \emph{Advances in neural information processing systems}, 5, 1992.

\bibitem[Benbaki et~al.(2023)Benbaki, Chen, Meng, Hazimeh, Ponomareva, Zhao,
  and Mazumder]{benbaki2023fast}
Riade Benbaki, Wenyu Chen, Xiang Meng, Hussein Hazimeh, Natalia Ponomareva, Zhe
  Zhao, and Rahul Mazumder.
\newblock Fast as chita: Neural network pruning with combinatorial
  optimization.
\newblock \emph{arXiv preprint arXiv:2302.14623}, 2023.

\bibitem[Lebedev and Lempitsky(2016)]{lebedev2016fast}
Vadim Lebedev and Victor Lempitsky.
\newblock Fast convnets using group-wise brain damage.
\newblock In \emph{Proceedings of the IEEE conference on computer vision and
  pattern recognition}, pages 2554--2564, 2016.

\bibitem[Wen et~al.(2016)Wen, Wu, Wang, Chen, and Li]{wen2016learning}
Wei Wen, Chunpeng Wu, Yandan Wang, Yiran Chen, and Hai Li.
\newblock Learning structured sparsity in deep neural networks.
\newblock \emph{Advances in neural information processing systems}, 29, 2016.

\bibitem[Voita et~al.(2019)Voita, Talbot, Moiseev, Sennrich, and
  Titov]{voita2019analyzing}
Elena Voita, David Talbot, Fedor Moiseev, Rico Sennrich, and Ivan Titov.
\newblock Analyzing multi-head self-attention: Specialized heads do the heavy
  lifting, the rest can be pruned.
\newblock \emph{arXiv preprint arXiv:1905.09418}, 2019.

\bibitem[El~Halabi et~al.(2022)El~Halabi, Srinivas, and
  Lacoste-Julien]{el2022data}
Marwa El~Halabi, Suraj Srinivas, and Simon Lacoste-Julien.
\newblock Data-efficient structured pruning via submodular optimization.
\newblock \emph{Advances in Neural Information Processing Systems},
  35:\penalty0 36613--36626, 2022.

\bibitem[Zhou et~al.(2021)Zhou, Ma, Zhu, Liu, Zhang, Yuan, Sun, and
  Li]{zhou2021learning}
Aojun Zhou, Yukun Ma, Junnan Zhu, Jianbo Liu, Zhijie Zhang, Kun Yuan, Wenxiu
  Sun, and Hongsheng Li.
\newblock Learning n: m fine-grained structured sparse neural networks from
  scratch.
\newblock \emph{arXiv preprint arXiv:2102.04010}, 2021.

\bibitem[Han et~al.(2015)Han, Pool, Tran, and Dally]{han2015learning}
Song Han, Jeff Pool, John Tran, and William Dally.
\newblock Learning both weights and connections for efficient neural network.
\newblock \emph{Advances in neural information processing systems}, 28, 2015.

\bibitem[Guo et~al.(2016)Guo, Yao, and Chen]{guo2016dynamic}
Yiwen Guo, Anbang Yao, and Yurong Chen.
\newblock Dynamic network surgery for efficient dnns.
\newblock \emph{Advances in neural information processing systems}, 29, 2016.

\bibitem[Kurtic et~al.(2022)Kurtic, Campos, Nguyen, Frantar, Kurtz, Fineran,
  Goin, and Alistarh]{kurtic2022optimal}
Eldar Kurtic, Daniel Campos, Tuan Nguyen, Elias Frantar, Mark Kurtz, Benjamin
  Fineran, Michael Goin, and Dan Alistarh.
\newblock The optimal bert surgeon: Scalable and accurate second-order pruning
  for large language models, 2022.
\newblock URL \url{https://arxiv.org/abs/2203.07259}.

\bibitem[Kurtz et~al.(2020)Kurtz, Kopinsky, Gelashvili, Matveev, Carr, Goin,
  Leiserson, Moore, Nell, Shavit, and Alistarh]{pmlr-v119-kurtz20a}
Mark Kurtz, Justin Kopinsky, Rati Gelashvili, Alexander Matveev, John Carr,
  Michael Goin, William Leiserson, Sage Moore, Bill Nell, Nir Shavit, and Dan
  Alistarh.
\newblock Inducing and exploiting activation sparsity for fast inference on
  deep neural networks.
\newblock In Hal~Daumé III and Aarti Singh, editors, \emph{Proceedings of the
  37th International Conference on Machine Learning}, volume 119 of
  \emph{Proceedings of Machine Learning Research}, pages 5533--5543, Virtual,
  13--18 Jul 2020. PMLR.
\newblock URL \url{http://proceedings.mlr.press/v119/kurtz20a.html}.

\bibitem[Iofinova et~al.(2021)Iofinova, Peste, Kurtz, and
  Alistarh]{DBLP:journals/corr/abs-2111-13445}
Eugenia Iofinova, Alexandra Peste, Mark Kurtz, and Dan Alistarh.
\newblock How well do sparse imagenet models transfer?
\newblock \emph{CoRR}, abs/2111.13445, 2021.
\newblock URL \url{https://arxiv.org/abs/2111.13445}.

\bibitem[Meng et~al.(2024{\natexlab{a}})Meng, Behdin, Wang, and
  Mazumder]{meng2024alps}
Xiang Meng, Kayhan Behdin, Haoyue Wang, and Rahul Mazumder.
\newblock Alps: Improved optimization for highly sparse one-shot pruning for
  large language models.
\newblock \emph{arXiv preprint arXiv:2406.07831}, 2024{\natexlab{a}}.

\bibitem[Frantar and Alistarh(2023)]{frantar2023sparsegpt}
Elias Frantar and Dan Alistarh.
\newblock Sparsegpt: Massive language models can be accurately pruned in
  one-shot.
\newblock In \emph{International Conference on Machine Learning}, pages
  10323--10337. PMLR, 2023.

\bibitem[Sun et~al.(2023)Sun, Liu, Bair, and Kolter]{sun2023simple}
Mingjie Sun, Zhuang Liu, Anna Bair, and J~Zico Kolter.
\newblock A simple and effective pruning approach for large language models.
\newblock \emph{arXiv preprint arXiv:2306.11695}, 2023.

\bibitem[Zhang et~al.(2023)Zhang, Zhao, Lin, Sun, Yao, Han, Tanner, Liu, and
  Ji]{zhang2023dynamic}
Yuxin Zhang, Lirui Zhao, Mingbao Lin, Yunyun Sun, Yiwu Yao, Xingjia Han, Jared
  Tanner, Shiwei Liu, and Rongrong Ji.
\newblock Dynamic sparse no training: Training-free fine-tuning for sparse
  llms.
\newblock \emph{arXiv preprint arXiv:2310.08915}, 2023.

\bibitem[Zhang et~al.(2022)Zhang, Roller, Goyal, Artetxe, Chen, Chen, Dewan,
  Diab, Li, Lin, et~al.]{opt}
Susan Zhang, Stephen Roller, Naman Goyal, Mikel Artetxe, Moya Chen, Shuohui
  Chen, Christopher Dewan, Mona Diab, Xian Li, Xi~Victoria Lin, et~al.
\newblock Opt: Open pre-trained transformer language models.
\newblock \emph{arXiv preprint arXiv:2205.01068}, 2022.

\bibitem[Kingma(2014)]{kingma2014adam}
Diederik~P Kingma.
\newblock Adam: A method for stochastic optimization.
\newblock \emph{arXiv preprint arXiv:1412.6980}, 2014.

\bibitem[Hinterm{\"u}ller and Wu(2015)]{hintermuller2015robust}
Michael Hinterm{\"u}ller and Tao Wu.
\newblock Robust principal component pursuit via inexact alternating
  minimization on matrix manifolds.
\newblock \emph{Journal of Mathematical Imaging and Vision}, 51\penalty0
  (3):\penalty0 361--377, 2015.

\bibitem[Cand{\`e}s et~al.(2011)Cand{\`e}s, Li, Ma, and
  Wright]{candes2011robust}
Emmanuel~J Cand{\`e}s, Xiaodong Li, Yi~Ma, and John Wright.
\newblock Robust principal component analysis?
\newblock \emph{Journal of the ACM (JACM)}, 58\penalty0 (3):\penalty0 1--37,
  2011.

\bibitem[Lin et~al.(2011)Lin, Liu, and Su]{lin2011linearized}
Zhouchen Lin, Risheng Liu, and Zhixun Su.
\newblock Linearized alternating direction method with adaptive penalty for
  low-rank representation.
\newblock \emph{Advances in neural information processing systems}, 24, 2011.

\bibitem[Chandrasekaran et~al.(2009)Chandrasekaran, Sanghavi, Parrilo, and
  Willsky]{5394889}
Venkat Chandrasekaran, Sujay Sanghavi, Pablo~A. Parrilo, and Alan~S. Willsky.
\newblock Sparse and low-rank matrix decompositions.
\newblock In \emph{2009 47th Annual Allerton Conference on Communication,
  Control, and Computing (Allerton)}, pages 962--967, 2009.
\newblock \doi{10.1109/ALLERTON.2009.5394889}.

\bibitem[Zhou and Tao(2011)]{zhou2011godec}
Tianyi Zhou and Dacheng Tao.
\newblock Godec: Randomized low-rank \& sparse matrix decomposition in noisy
  case.
\newblock In \emph{Proceedings of the 28th International Conference on Machine
  Learning, ICML 2011}, 2011.

\bibitem[Bertsimas et~al.(2023)Bertsimas, Cory-Wright, and
  Johnson]{JMLR:v24:21-1130}
Dimitris Bertsimas, Ryan Cory-Wright, and Nicholas A.~G. Johnson.
\newblock Sparse plus low rank matrix decomposition: A discrete optimization
  approach.
\newblock \emph{Journal of Machine Learning Research}, 24\penalty0
  (267):\penalty0 1--51, 2023.
\newblock URL \url{http://jmlr.org/papers/v24/21-1130.html}.

\bibitem[Netrapalli et~al.(2014)Netrapalli, U~N, Sanghavi, Anandkumar, and
  Jain]{NIPS2014_443cb001}
Praneeth Netrapalli, Niranjan U~N, Sujay Sanghavi, Animashree Anandkumar, and
  Prateek Jain.
\newblock Non-convex robust pca.
\newblock In Z.~Ghahramani, M.~Welling, C.~Cortes, N.~Lawrence, and K.Q.
  Weinberger, editors, \emph{Advances in Neural Information Processing
  Systems}, volume~27. Curran Associates, Inc., 2014.
\newblock URL
  \url{https://proceedings.neurips.cc/paper_files/paper/2014/file/443cb001c138b2561a0d90720d6ce111-Paper.pdf}.

\bibitem[Yu et~al.(2017)Yu, Liu, Wang, and Tao]{yu2017compressing}
Xiyu Yu, Tongliang Liu, Xinchao Wang, and Dacheng Tao.
\newblock On compressing deep models by low rank and sparse decomposition.
\newblock In \emph{Proceedings of the IEEE conference on computer vision and
  pattern recognition}, pages 7370--7379, 2017.

\bibitem[Li et~al.(2023{\natexlab{a}})Li, Yu, Zhang, Liang, He, Chen, and
  Zhao]{li2023losparse}
Yixiao Li, Yifan Yu, Qingru Zhang, Chen Liang, Pengcheng He, Weizhu Chen, and
  Tuo Zhao.
\newblock Losparse: Structured compression of large language models based on
  low-rank and sparse approximation.
\newblock In \emph{International Conference on Machine Learning}, pages
  20336--20350. PMLR, 2023{\natexlab{a}}.

\bibitem[Zhang and Papyan(2024)]{zhang2024oats}
Stephen Zhang and Vardan Papyan.
\newblock Oats: Outlier-aware pruning through sparse and low rank
  decomposition.
\newblock \emph{arXiv preprint arXiv:2409.13652}, 2024.

\bibitem[Mozaffari et~al.(2024)Mozaffari, Yazdanbakhsh, Zhang, and
  Dehnavi]{mozaffari2024slope}
Mohammad Mozaffari, Amir Yazdanbakhsh, Zhao Zhang, and Maryam~Mehri Dehnavi.
\newblock Slope: Double-pruned sparse plus lazy low-rank adapter pretraining of
  llms.
\newblock \emph{arXiv preprint arXiv:2405.16325}, 2024.

\bibitem[Sze et~al.(2020)Sze, Chen, Yang, and Emer]{sze2020efficient}
Vivienne Sze, Yu-Hsin Chen, Tien-Ju Yang, and Joel~S Emer.
\newblock \emph{Efficient processing of deep neural networks}.
\newblock Springer, 2020.

\bibitem[Li et~al.(2023{\natexlab{b}})Li, Yu, Liang, He, Karampatziakis, Chen,
  and Zhao]{li2023loftq}
Yixiao Li, Yifan Yu, Chen Liang, Pengcheng He, Nikos Karampatziakis, Weizhu
  Chen, and Tuo Zhao.
\newblock Loftq: Lora-fine-tuning-aware quantization for large language models.
\newblock \emph{arXiv preprint arXiv:2310.08659}, 2023{\natexlab{b}}.

\bibitem[Guo et~al.(2023)Guo, Greengard, Xing, and Kim]{guo2023lq}
Han Guo, Philip Greengard, Eric~P Xing, and Yoon Kim.
\newblock Lq-lora: Low-rank plus quantized matrix decomposition for efficient
  language model finetuning.
\newblock \emph{arXiv preprint arXiv:2311.12023}, 2023.

\bibitem[Eckart and Young(1936)]{eckart1936approximation}
Carl Eckart and Gale Young.
\newblock The approximation of one matrix by another of lower rank.
\newblock \emph{Psychometrika}, 1\penalty0 (3):\penalty0 211--218, 1936.

\bibitem[Chandrasekaran et~al.(2011)Chandrasekaran, Sanghavi, Parrilo, and
  Willsky]{chandrasekaran2011rank}
Venkat Chandrasekaran, Sujay Sanghavi, Pablo~A Parrilo, and Alan~S Willsky.
\newblock Rank-sparsity incoherence for matrix decomposition.
\newblock \emph{SIAM Journal on Optimization}, 21\penalty0 (2):\penalty0
  572--596, 2011.

\bibitem[Boyd et~al.(2011)Boyd, Parikh, Chu, Peleato, Eckstein,
  et~al.]{boyd2011distributed}
Stephen Boyd, Neal Parikh, Eric Chu, Borja Peleato, Jonathan Eckstein, et~al.
\newblock Distributed optimization and statistical learning via the alternating
  direction method of multipliers.
\newblock \emph{Foundations and Trends{\textregistered} in Machine learning},
  3\penalty0 (1):\penalty0 1--122, 2011.

\bibitem[Meng et~al.(2024{\natexlab{b}})Meng, Ibrahim, Behdin, Hazimeh,
  Ponomareva, and Mazumder]{meng2024osscar}
Xiang Meng, Shibal Ibrahim, Kayhan Behdin, Hussein Hazimeh, Natalia Ponomareva,
  and Rahul Mazumder.
\newblock Osscar: One-shot structured pruning in vision and language models
  with combinatorial optimization.
\newblock \emph{arXiv preprint arXiv:2403.12983}, 2024{\natexlab{b}}.

\bibitem[Halko et~al.(2011)Halko, Martinsson, and Tropp]{halko2011finding}
Nathan Halko, Per-Gunnar Martinsson, and Joel~A Tropp.
\newblock Finding structure with randomness: Probabilistic algorithms for
  constructing approximate matrix decompositions.
\newblock \emph{SIAM review}, 53\penalty0 (2):\penalty0 217--288, 2011.

\bibitem[Raffel et~al.(2020)Raffel, Shazeer, Roberts, Lee, Narang, Matena,
  Zhou, Li, and Liu]{JMLR:v21:20-074}
Colin Raffel, Noam Shazeer, Adam Roberts, Katherine Lee, Sharan Narang, Michael
  Matena, Yanqi Zhou, Wei Li, and Peter~J. Liu.
\newblock Exploring the limits of transfer learning with a unified text-to-text
  transformer.
\newblock \emph{Journal of Machine Learning Research}, 21\penalty0
  (140):\penalty0 1--67, 2020.
\newblock URL \url{http://jmlr.org/papers/v21/20-074.html}.

\bibitem[Per(2022)]{Perplexity}
Perplexity of fixed-length models, 2022.
\newblock URL \url{https://huggingface.co/docs/transformers/perplexity}.

\bibitem[Merity et~al.(2017)Merity, Xiong, Bradbury, and
  Socher]{merity2017pointer}
Stephen Merity, Caiming Xiong, James Bradbury, and Richard Socher.
\newblock Pointer sentinel mixture models.
\newblock In \emph{International Conference on Learning Representations}, 2017.
\newblock URL \url{https://openreview.net/forum?id=Byj72udxe}.

\bibitem[Marcus et~al.(1994)Marcus, Kim, Marcinkiewicz, MacIntyre, Bies,
  Ferguson, Katz, and Schasberger]{Marcus1994}
Mitchell Marcus, Grace Kim, Mary~Ann Marcinkiewicz, Robert MacIntyre, Ann Bies,
  Mark Ferguson, Karen Katz, and Britta Schasberger.
\newblock The penn treebank: Annotating predicate argument structure.
\newblock In \emph{Proceedings of the Workshop on Human Language Technology},
  HLT '94, page 114–119, USA, 1994. Association for Computational
  Linguistics.
\newblock ISBN 1558603573.
\newblock \doi{10.3115/1075812.1075835}.
\newblock URL \url{https://doi.org/10.3115/1075812.1075835}.

\bibitem[Gao et~al.()Gao, Tow, Abbasi, Biderman, Black, DiPofi, Foster,
  Golding, Hsu, Le~Noac’h, et~al.]{gao10256836framework}
L~Gao, J~Tow, B~Abbasi, S~Biderman, S~Black, A~DiPofi, C~Foster, L~Golding,
  J~Hsu, A~Le~Noac’h, et~al.
\newblock A framework for few-shot language model evaluation, 12 2023.
\newblock \emph{URL https://zenodo. org/records/10256836}, 7.

\bibitem[Bisk et~al.(2020)Bisk, Zellers, Gao, Choi, et~al.]{bisk2020piqa}
Yonatan Bisk, Rowan Zellers, Jianfeng Gao, Yejin Choi, et~al.
\newblock Piqa: Reasoning about physical commonsense in natural language.
\newblock In \emph{Proceedings of the AAAI conference on artificial
  intelligence}, volume~34, pages 7432--7439, 2020.

\bibitem[Clark et~al.(2018)Clark, Cowhey, Etzioni, Khot, Sabharwal, Schoenick,
  and Tafjord]{clark2018think}
Peter Clark, Isaac Cowhey, Oren Etzioni, Tushar Khot, Ashish Sabharwal, Carissa
  Schoenick, and Oyvind Tafjord.
\newblock Think you have solved question answering? try arc, the ai2 reasoning
  challenge.
\newblock \emph{arXiv preprint arXiv:1803.05457}, 2018.

\bibitem[Zellers et~al.(2019)Zellers, Holtzman, Bisk, Farhadi, and
  Choi]{zellers2019hellaswag}
Rowan Zellers, Ari Holtzman, Yonatan Bisk, Ali Farhadi, and Yejin Choi.
\newblock Hellaswag: Can a machine really finish your sentence?
\newblock \emph{arXiv preprint arXiv:1905.07830}, 2019.

\bibitem[Sakaguchi et~al.(2021)Sakaguchi, Bras, Bhagavatula, and
  Choi]{sakaguchi2021winogrande}
Keisuke Sakaguchi, Ronan~Le Bras, Chandra Bhagavatula, and Yejin Choi.
\newblock Winogrande: An adversarial winograd schema challenge at scale.
\newblock \emph{Communications of the ACM}, 64\penalty0 (9):\penalty0 99--106,
  2021.

\bibitem[Poliak(2020)]{poliak2020survey}
Adam Poliak.
\newblock A survey on recognizing textual entailment as an nlp evaluation.
\newblock \emph{arXiv preprint arXiv:2010.03061}, 2020.

\bibitem[Banerjee et~al.(2019)Banerjee, Pal, Mitra, and
  Baral]{banerjee2019careful}
Pratyay Banerjee, Kuntal~Kumar Pal, Arindam Mitra, and Chitta Baral.
\newblock Careful selection of knowledge to solve open book question answering.
\newblock \emph{arXiv preprint arXiv:1907.10738}, 2019.

\bibitem[Clark et~al.(2019)Clark, Lee, Chang, Kwiatkowski, Collins, and
  Toutanova]{clark2019boolq}
Christopher Clark, Kenton Lee, Ming-Wei Chang, Tom Kwiatkowski, Michael
  Collins, and Kristina Toutanova.
\newblock Boolq: Exploring the surprising difficulty of natural yes/no
  questions.
\newblock \emph{arXiv preprint arXiv:1905.10044}, 2019.

\bibitem[Reuther et~al.(2018)Reuther, Kepner, Byun, Samsi, Arcand, Bestor,
  Bergeron, Gadepally, Houle, Hubbell, Jones, Klein, Milechin, Mullen, Prout,
  Rosa, Yee, and Michaleas]{reuther2018interactive}
Albert Reuther, Jeremy Kepner, Chansup Byun, Siddharth Samsi, William Arcand,
  David Bestor, Bill Bergeron, Vijay Gadepally, Michael Houle, Matthew Hubbell,
  Michael Jones, Anna Klein, Lauren Milechin, Julia Mullen, Andrew Prout,
  Antonio Rosa, Charles Yee, and Peter Michaleas.
\newblock Interactive supercomputing on 40,000 cores for machine learning and
  data analysis.
\newblock In \emph{2018 IEEE High Performance extreme Computing Conference
  (HPEC)}, page 1–6. IEEE, 2018.

\end{thebibliography}
\clearpage

\appendix

\section{Experimental Details}
\label{supp:experiments-supp}


\subsection{Experimental Setup}
Following the framework proposed by \citet{frantar2023sparsegpt} for one-shot pruning, we minimize \cref{eq:matrix-decomposition} sequentially, layer by layer. For a given layer $\ell$, the input activation matrix $\bfX$ introduced in \cref{sec:optimization-formulation} is the output of the previous $\ell - 1$ compressed layers (sparse plus low-rank) using $N$ calibration samples. 

\textbf{Implementation details.}
\begin{itemize}
    \item For the construction of the Hessian matrix $\bfH = \bfXTX$ introduced in \cref{sec:algorithm-design}, we use the same setup of SparseGPT \cite{frantar2023sparsegpt} and we use the author's implementation of SparseGPT---as a pruning plug-in method to minimize \Pone (codes available on GitHub).
    \item We utilize the author's implementation of OATS \cite{zhang2024oats} with the default hyperparameter settings to show LLM evaluation benchmarks and layer-wise reconstruction error in \cref{fig:reconstruction-error}.
    \item The LLM evaluation benchmarks reported in \cref{tab:slr-fixed-compression} are retrieved from the paper ALPS by \citet{meng2024alps} which uses the same evaluation strategy (and code) we do for the reported tasks [other zero-shot tasks are not reported in ALPS]. We report all zero-shot tasks results for OATS and \ourframework in \cref{tab:supp-slr-fixed-compression}.
\end{itemize}

\subsection{Hyperparameter Choice}
The hyperparameters used in \ourframework are the following: $\lambda = 0.01 \Tr{\bfH}$; default value in SparseGPT. $T_\text{AM}$ is set to be 80; default value in OATS. $T_\text{LR} = 50$; we propose this default value for all experiments. $\eta = 1e^{-2}$; we propose this default value for all experiments (only works well with the scaling introduced in \cref{scaling-low-rank}). $r$ is either set to $64$ and fixed for all layers, or is flexible and given by the formula $r = \left\lfloor {(1 - \rho - \frac{N}{M}) \cdot(\Nout \cdot \Nin)} / \pr{\Nout + \Nin} \right\rfloor$ introduced in \cref{sec:experimental-results}. \texttt{Prune}; we propose by default to use SparseGPT. \texttt{Optimizer}; we propose the Adam optimizer. \textbf{is\_scaled}; we propose to set this to True by default. It converges faster in practice and allows to skip the tuning of the learning rate $\eta$.

\subsection{Additional Experimental Resuls}
\textbf{N:M Sparsity + Fixed Compression Ratio:}
This is the same setting described in \cref{sec:experimental-results}. We extend the results reported in \cref{tab:slr-fixed-compression} to include the 8 zero-shot tasks and the Llama3.2 model. Results are reported in \cref{tab:supp-slr-fixed-compression}.
\begin{table}[h!]
\centering
\resizebox{1.0\textwidth}{!}{%
\renewcommand{\arraystretch}{1.3}
\begin{tabular}{cccccccccccccc}
\toprule
\multirow{2.25}{*}{\textbf{Model}} & \multirow{2.25}{*}{\textbf{Algorithm}} & \multicolumn{3}{c}{\textbf{Perplexity ($\downarrow$)}} & \multicolumn{9}{c}{\textbf{Zero-shot ($\uparrow$)}} \\ 
\cmidrule(rl){3-5} \cmidrule(rl){6-14}
& & \textbf{C4} & \textbf{WT2} & \textbf{PTB} & \textbf{PIQA} & \textbf{HS} & \textbf{ARC-E} & \textbf{ARC-C} & \textbf{WG} & \textbf{RTE} & \textbf{OQA} & \textbf{BoolQ} & \textbf{Avg}\\ 
\midrule
\multirow{5.5}{*}{Llama3-8B} 
&\texttt{OATS-2:8+LR}       & 21.03 & \textbf{14.54} & 24.15 & 73.67 & \textbf{62.42} & 59.68 & \textbf{37.12} & 65.43 & 55.23 & \textbf{36.40} & 73.98 & 57.99 \\
&\texttt{Ours-2:8+LR}       & \textbf{20.05} & 15.03 & \textbf{22.01} & \textbf{74.05} & 60.69 & \textbf{60.52} & 36.18 & \textbf{66.77} & \textbf{57.04} & 35.00 & \textbf{76.02} & \textbf{58.28} \\
\cmidrule(rl){2-2}
&\texttt{OATS-3:8+LR}       & 16.87 & 11.43 & 18.53 & 75.24 & 66.90 & 65.91 & 39.85 & 68.90 & 61.37 & 39.00 & 76.61 & 61.72 \\
&\texttt{Ours-3:8+LR}       & \textbf{16.16} & \textbf{11.36} & \textbf{16.71} & \textbf{75.79} & \textbf{67.33} & \textbf{67.55} & \textbf{41.04} & \textbf{69.53} & \textbf{58.48} & \textbf{39.20} & \textbf{79.91} & \textbf{62.35} \\
\cmidrule(rl){2-2}
&\texttt{dense}               & 9.44 & 6.14 & 11.18 & 80.79 & 79.17 & 77.69 & 53.33 & 72.85 & 69.68 & 45.00 & 81.44 & 69.99 \\

\midrule
\multirow{5.5}{*}{Llama3.2-1B} 
&\texttt{OATS-2:8+LR}       & 78.18 & 53.05 & 80.17 & 59.03 & 36.42 & 37.08 & 22.87 & 52.80 & 52.71 & 27.40 & 61.77 & 43.76 \\
&\texttt{Ours-2:8+LR}       & \textbf{41.08} & \textbf{30.92} & \textbf{48.85} & \textbf{63.22} & \textbf{39.07} & \textbf{42.55} & \textbf{25.77} & \textbf{55.17} & \textbf{53.07} & \textbf{28.00} & \textbf{62.11} & \textbf{46.12} \\
\cmidrule(rl){2-2}
&\texttt{OATS-3:8+LR}       & 42.81 & 29.35 & 47.58 & 63.49 & 42.25 & 43.43 & 25.09 & 54.85 & 52.35 & \textbf{29.60} & 62.05 & 46.64 \\
&\texttt{Ours-3:8+LR}       & \textbf{31.35} & \textbf{22.89} & \textbf{34.99} & \textbf{66.43} & \textbf{45.00} & \textbf{46.42} & \textbf{25.85} & \textbf{56.43} & \textbf{52.71} & 28.80 & \textbf{62.26} & \textbf{47.99} \\
\cmidrule(rl){2-2}
&\texttt{dense}               & 14.01 & 9.75 & 17.59 & 74.59 & 63.66 & 60.48 & 36.26 & 60.69 & 56.68 & 37.20 & 63.98 & 56.69 \\

\midrule
\multirow{5.5}{*}{Llama3.2-3B} 
&\texttt{OATS-2:8+LR}       & 30.73 & 22.65 & 36.31 & 68.55 & 51.76 & 54.46 & \textbf{31.14} & 61.17 & \textbf{58.48} & \textbf{30.80} & \textbf{70.43} & \textbf{53.35} \\
&\texttt{Ours-2:8+LR}       & \textbf{25.22} & \textbf{19.61} & \textbf{29.54} & \textbf{69.59} & \textbf{52.94} & \textbf{55.30} & 29.69 & \textbf{62.67} & 55.23 & 30.60 & 69.24 & 53.16\\
\cmidrule(rl){2-2}
&\texttt{OATS-3:8+LR}       & 21.96 & 15.84 & 26.22 & \textbf{72.69} & 58.61 & \textbf{58.92} & \textbf{34.13} & 63.14 & \textbf{58.12} & 33.60 & 67.22 & 55.80 \\
&\texttt{Ours-3:8+LR}       & \textbf{20.03} & \textbf{14.85} & \textbf{22.92} & 72.42 & \textbf{58.92} & 56.69 & 33.53 & \textbf{64.01} & 56.32 & \textbf{37.00} & \textbf{70.31} & \textbf{56.15} \\
\cmidrule(rl){2-2}
&\texttt{dense}               & 11.33 & 7.81 & 13.53 & 77.48 & 73.61 & 71.63 & 45.99 & 69.85 & 54.51 & 43.00 & 73.39 & 63.68 \\

\bottomrule
\end{tabular}
}
\vspace{3pt}
\caption{Performance analysis for one-shot N:M sparse plus a low-rank matrix decomposition of Llama3 and Llama3.2 models. The compression ratio is fixed to be $\rho=0.5$. For Perplexity, $(\downarrow)$ lower values are preferred. For zero-shot tasks, $(\uparrow)$ higher values are preferred.}
\label{tab:supp-slr-fixed-compression}
\end{table}

\textbf{Unstructured Sparsity + Fixed Rank Ratio:} This is the setting introduced in OATS \cite{zhang2024oats}. This scheme takes as inputs a compression ratio $\rho$ (e.g. $50\%$) and rank ratio $\kappa$ (e.g. $0.3$; default value in OATS for the Llama3-8B model). The rank of the low-rank component $r$ and the number of non-zeros $k$ in the unstructured sparsity are given by.
\begin{equation*}
    r = \left\lfloor \kappa \cdot (1 - \rho) \cdot \frac{\Nout \cdot \Nin}{\Nout + \Nin} \right\rfloor, \quad \quad k = \left\lfloor (1 - \kappa) \cdot (1 - \rho) \cdot \Nout \cdot \Nin \right\rfloor.
\end{equation*}
See OATS for a discussion on how to choose the rank ratio $\kappa$ for a given model. Note that OATS introduces OWL ratios--different sparsity budgets for different layers to reduce the utility drop. The results for this setting do not apply OWL and consider uniform unstructured sparsity throughout layers. Results for OATS and \ourframework are reported in \cref{tab:supp-slr-unstructured}.

\begin{table}[h!]
\centering
\resizebox{1.0\textwidth}{!}{%
\renewcommand{\arraystretch}{1.3}
\begin{tabular}{ccc@{\hskip 8pt}cccc@{\hskip 8pt}ccccccccc}
\toprule
\multirow{2.25}{*}{\textbf{Model}} & \multirow{2.25}{*}{\textbf{Algorithm}} && \multicolumn{3}{c}{\textbf{Perplexity ($\downarrow$)}} && \multicolumn{9}{c}{\textbf{Zero-shot ($\uparrow$)}} \\ 
\cmidrule(rl){4-6} \cmidrule(r){8-16}
&&& \textbf{C4} & \textbf{WT2} & \textbf{PTB} && \textbf{PIQA} & \textbf{HS} & \textbf{ARC-E} & \textbf{ARC-C} & \textbf{WG} & \textbf{RTE} & \textbf{OQA} & \textbf{BoolQ} & \textbf{Avg}\\ 

\midrule
\multirow{7.5}{*}{Llama3-8B}
&\texttt{OATS-60\%+LR}       && 23.61 & 16.52 & 25.85 && 72.91 & 59.65 & \textbf{60.10} & 33.36 & 65.35 & 53.07 & 31.60 & \textbf{75.96} & 56.50 \\
&\texttt{Ours-60\%+LR}       &&  \textbf{20.70} & \textbf{15.66} & \textbf{23.31} && \textbf{73.29} & \textbf{60.58} & 59.26 & \textbf{34.64} & \textbf{67.88} & \textbf{53.43} & \textbf{35.40} & 75.08 & \textbf{57.44} \\
\cmidrule(rl){2-2}
&\texttt{OATS-70\%+LR}       &&  106.98 & 81.77 & 110.44 && 55.60 & 30.30 & 32.45 & 20.05 & 49.96 & \textbf{52.71} & 27.00 & 62.35 & 41.30 \\
&\texttt{Ours-70\%+LR}       && \textbf{50.07} & \textbf{49.13} & \textbf{60.89} && \textbf{60.50} & \textbf{39.67} & \textbf{37.21} & \textbf{23.38} & \textbf{55.25} & \textbf{52.71} & \textbf{27.40} & \textbf{66.09} & \textbf{45.27} \\

\cmidrule(rl){2-2}
&\texttt{OATS-80\%+LR}       && 748.40 & 909.75 & 1601.02 && 52.29 & 27.25 & 26.81 & \textbf{24.40} & 47.59 & \textbf{52.71} & \textbf{26.60} & 37.83 & 36.93 \\
&\texttt{Ours-80\%+LR}       && \textbf{164.27} & \textbf{265.28} & \textbf{235.38} && \textbf{53.32} & \textbf{28.53} & \textbf{29.38} & 20.22 & \textbf{49.49} & \textbf{52.71} & \textbf{26.60} & \textbf{38.84} & \textbf{37.39} \\
\cmidrule(rl){2-2}
&\texttt{dense}               && 11.33 & 7.81 & 13.53 && 77.48 & 73.61 & 71.63 & 45.99 & 69.85 & 54.51 & 43.00 & 73.39 & 63.68 \\

\midrule
\multirow{7.5}{*}{Llama3.2-3B} 
&\texttt{OATS-60\%+LR}       && 34.57 & 24.94 & 41.51 && 67.79 & 48.40 & \textbf{52.57} & \textbf{30.38} & 57.70 & 54.15 & \textbf{30.80} & 65.66 & 50.93 \\
&\texttt{Ours-60\%+LR}       && \textbf{27.67} & \textbf{21.90} & \textbf{33.40} && \textbf{69.15} & \textbf{52.04} & 51.26 & 29.52 & \textbf{61.96} & \textbf{58.12} & 29.80 & \textbf{69.72} & \textbf{52.70} \\
\cmidrule(rl){2-2}
&\texttt{OATS-70\%+LR}       &&  155.48 & 121.76 & 167.60 && 54.57 & 29.83 & 30.43 & 21.42 & \textbf{49.64} & \textbf{52.71} & \textbf{28.20} & 60.43 & 40.90 \\
&\texttt{Ours-70\%+LR}       && \textbf{78.65} & \textbf{75.23} & \textbf{103.10} && \textbf{58.43} & \textbf{32.44} & \textbf{35.27} & \textbf{21.67} & 49.41 & \textbf{52.71} & 27.00 & \textbf{62.29} & \textbf{42.40} \\
\cmidrule(rl){2-2}
&\texttt{OATS-80\%+LR}       && 1085.27 & 1610.87 & 2546.29 && 50.60 & 26.60 & 26.68 & \textbf{24.40} & 47.67 & \textbf{52.71} & \textbf{26.60} & 37.83 & 36.64 \\
&\texttt{Ours-80\%+LR}       && \textbf{217.62} & \textbf{320.98} & \textbf{320.02} && \textbf{53.10} & \textbf{27.86} & \textbf{29.12} & 22.01 & \textbf{47.75} & 50.54 & \textbf{26.60} & \textbf{46.61} & \textbf{37.95} \\
\cmidrule(rl){2-2}
&\texttt{dense}               && 11.33 & 7.81 & 13.53 && 77.48 & 73.61 & 71.63 & 45.99 & 69.85 & 54.51 & 43.00 & 73.39 & 63.68 \\

\bottomrule
\end{tabular}
}
\vspace{2pt}
\caption{Performance analysis for one-shot unstructured sparsity plus a low-rank matrix decomposition of Llama3 and Llama3.2-3B model. The rank ratio of the low-rank component is fixed to be $\kappa=0.3$. For Perplexity, $(\downarrow)$ lower values are preferred. For zero-shot tasks, $(\uparrow)$ higher values are preferred.}
\label{tab:supp-slr-unstructured}
\end{table}

\end{document}